\documentclass[11pt,a4paper]{article}
\usepackage[hyperref]{emnlp-ijcnlp-2019}
\usepackage{times}
\usepackage{latexsym}
\usepackage{adjustbox}
\usepackage{blindtext}
\usepackage{graphicx}
\usepackage{subcaption}
\usepackage{listings}
\usepackage{footnote}
\usepackage{url}
\usepackage{float}
\usepackage{capt-of}

\aclfinalcopy 

\title{LIDA: Lightweight Interactive Dialogue Annotator}

\author{Edward Collins \\
  Wluper Ltd. \\
  London, United Kingdom \\
  \texttt{ed@wluper.com} \\\And 
  Nikolai Rozanov \\
  Wluper Ltd. \\
  London, United Kingdom\\
  \texttt{nikolai@wluper.com} \\\And
  Bingbing Zhang \\
  Wluper Ltd. \\
  London, United Kingdom \\
   \texttt{bingbing@wluper.com} \\
 }

\date{}

\begin{document}
\maketitle

\begin{abstract}

Dialogue systems have the potential to change how people interact with machines but are highly dependent on the quality of the data used to train them. It is therefore important to develop good dialogue annotation tools which can improve the speed and quality of dialogue data annotation. With this in mind, we introduce LIDA, an annotation tool designed specifically for conversation data. As far as we know, LIDA is the first dialogue annotation system that handles the entire dialogue annotation pipeline from raw text, as may be the output of transcription services, to structured conversation data. Furthermore it supports the integration of arbitrary machine learning models as annotation recommenders and also has a dedicated interface to resolve inter-annotator disagreements such as after crowdsourcing annotations for a dataset. LIDA is fully open source, documented and publicly available \footnote{\url{https://github.com/Wluper/lida}}.

\end{abstract}
\begin{table*}[ht]
\centering
\tiny
\begin{tabular}{p{0.20\linewidth}p{0.12\linewidth}p{0.12\linewidth}p{0.12\linewidth}p{0.12\linewidth}p{0.10\linewidth}p{0.05\linewidth}}
\hline
Annotation Tool & Turn/Dialogue Segmentation &  Classification Labels & Edit Dialogues/Turns & Recommenders & Inter-Annotator Disagreement Resolution & Language\\
\hline
LIDA                              & YES & YES & YES & YES & YES & PYTHON   \\
INCEpTion \cite{klie2018inception} & NO  & YES & NO  & YES & YES/NO\footnotemark[4] & JAVA     \\
GATE \cite{cunningham2002gate}     & NO  & YES & NO  & NO  & YES/NO \footnotemark[5]   & JAVA     \\
TWIST \cite{TWIST}                 & YES & NO  & YES & NO  & NO  & - \\
BRAT \cite{stenetorp2012brat}      & NO  & YES & NO  & YES & NO & PYTHON   \\
DOCCANO\footnotemark[3]            & NO  & YES & NO  & NO & NO & PYTHON   \\
DialogueView \cite{heeman-etal-2002-dialogueview}  & YES  & YES & YES  & NO & NO & TcK/TK \\
\hline
\end{tabular}
\caption{Annotator Tool Comparison Table}
\label{tab:tool_comparison}

\centering
\small
Turn/Dialogue Segmentation: segment raw text into turns and dialogues. Classification Labels: label classification data. Edit Dialogues/Turns: allow users to add/edit/delete new turns or dialogues. Recommenders: ML models to suggest annotations. Inter-Annotator Disagreement Resolution: whether the system has an interface to resolve disagreements between different annotators. Language: what programming language the system uses

\end{table*}

\section{Introduction}
Of all the milestones on the road to creating artificial general intelligence perhaps one of the most significant is giving machines the ability to converse with humans. Dialogue systems are becoming one of the most active research areas in Natural Language Processing (NLP) and Machine Learning (ML). New, large dialogue datasets such as MultiWOZ \cite{budzianowski2018multiwoz} have allowed data-hungry deep learning algorithms to be applied to dialogue systems, and challenges such as the Dialogue State Tracking Challenge (DSTC) \cite{henderson2014second} and Amazon's Alexa Prize \cite{alexaPrize} encourage competition among teams to produce the best systems.

The quality of a dialogue system dependents on the quality of the data used to train the system. Creating a high-quality dialogue dataset incurs a large annotation cost, which makes good dialogue annotation tools essential to ensure the highest possible quality. Many annotation tools exist for a range of NLP tasks but none are designed specifically for dialogue with modern usability principles in mind - in collecting MultiWOZ, for example, \newcite{budzianowski2018multiwoz} had to create a bespoke annotation interface.

In this paper, we introduce LIDA, a web application designed to make dialogue dataset creation and annotation as easy and fast as possible. In addition to following modern principles of usability, LIDA integrates best practices from other state-of-the-art annotation tools such as INCEpTION \cite{klie2018inception}, most importantly by allowing arbitrary ML models to be integrated as annotation recommenders to suggest annotations for data. Any system with the correct API can be integrated into LIDA's back end, meaning LIDA can be used as a front end for researchers to interact with their dialogue systems and correct their responses, then save the interaction as a future test case. 

When data is crowdsourced, it is good practice to have multiple annotators label each piece of data to reduce noise and mislabelling \cite{imagenet}. Once you have multiple annotations, it is important to be able to resolve conflicts by highlighting where annotators disagreed so that an arbiter can decide on the correct annotation. To this end, LIDA provides a dedicated interface which automatically finds where annotators have disagreed and displays the labels alongside a percentage of how many annotators selected each label, with the majority annotated labels selected by default.


\subsection{Main Contributions}
Our main contributions with this tool are:
\begin{itemize}
    \item A modern annotation tool designed specifically for task-oriented conversation data 
    \item The first dialogue annotator capable of handling the full dialogue annotation pipeline from turn and dialogue segmentation through to labelling structured conversation data 
    \item Easy integration of dialogue systems and recommenders to provide annotation suggestions
    \item A dedicated interface to resolve inter-annotator disagreements for dialogue data
\end{itemize}

\section{Related Work}

\par

Various annotation tools have been developed for NLP tasks in recent years. Table \ref{tab:tool_comparison} compares LIDA with other recent annotation tools. TWIST \cite{TWIST} is a dialogue annotation tool which consists of two stages: turn segmentation and content feature annotation. Turn segmentation allows users to highlight and create new turn segments from raw text. After this, users can annotate sections of text in a segment by highlighting them and selecting from a predefined feature list. However, this tool doesn't allow users to specify custom annotations or labels and doesn't support classification or slot-value annotation. This is not compatible with modern dialogue datasets which require such annotations \cite{budzianowski2018multiwoz}.

INCEpTION \cite{klie2018inception} is a semantic annotation platform for interactive tasks that require semantic resources like entity linking. It provides machine learning models to suggest annotations and allows users to collect and model knowledge directly in the tool. GATE \cite{cunningham2002gate} is an open source tool that provides predefined solutions for many text processing tasks. It is powerful because it allows annotators to enhance the provided annotation tools with their own Java code, making it easily extensible and provides an enormous number of predefined features. However, GATE is a large and complicated tool with a significant setup cost - its instruction manual alone is over 600 pages long\footnote{https://gate.ac.uk/sale/tao/tao.pdf}. Despite their large feature sets, INCEpTION and GATE are not designed for annotating dialogue and cannot display data as turns, an important feature for dialogue datasets. 


BRAT \cite{stenetorp2012brat} and Doccano\footnote{https://github.com/chakki-works/doccano} are web-based annotation tools for tasks such as text classification and sequence labeling. They have intuitive and user-friendly interfaces which aim to make the creation of certain types of dataset such as classification or sequence labelling datasets as fast as possible. BRAT also supports annotation suggestions by integrating ML models. However, like INCEpTION\footnote{Getting the scores is available as a plugin: https://dkpro.github.io/dkpro-statistics/dkpro-agreement-poster.pdf - resolving the issues seems to be not supported} and GATE\footnote{Again inter-annotator score calculation capabilities are available as separate plug-in https://gate.ac.uk/releases/gate-5.1-beta1-build3397-ALL/doc/tao/splitch10.html - however support for resolutions is not apparent}, they are not designed for annotating dialogues and do not support generation of formatted conversational data from a raw text file such as may be output by a transcription service. LIDA aims to fill these gaps by providing a lightweight, easy-to-setup annotation tool which displays data as a series of dialogues, supports integration of arbitrary ML models as recommenders and supports segmentation of raw text into dialogues and turns.


DialogueView \cite{heeman-etal-2002-dialogueview} is a tool for dialogue annotation. However, the main use-cases are not focused on building dialogue systems, rather it is focused on segmenting recorded conversations. It supports annotating audio files as well as discourse segmentation - hence, granular labelling of the dialogue, recommenders, inter-annotator agreement, and slot-value labelling is not possible with DialogueView.

\section{System Overview}

LIDA is built according to a client-server architecture with the front end written in standard web languages (HTML/CSS/JavaScript) that will run on any browser. The back end written in Python using the Flask\footnote{http://flask.pocoo.org/} web framework as a RESTful API. 

\begin{figure*}
  \centering
  \includegraphics[width=0.85\textwidth]{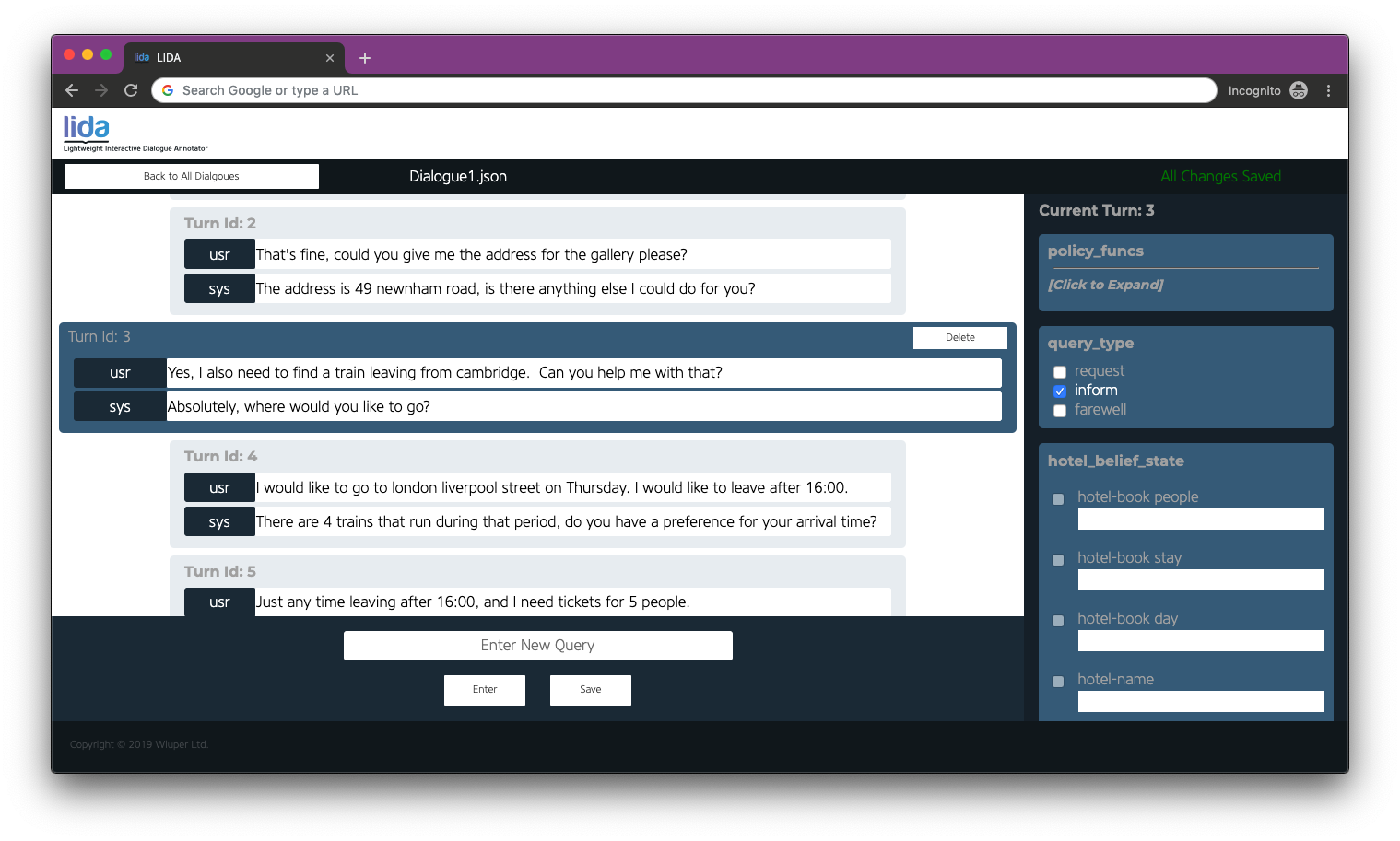}
  \caption{Turn List : A list of turns for one specific dialogue, users can add new turns, delete turns, edit utterances and annotate labels here.}
  \label{fig:turn_list}
\end{figure*}

The main screen which lists all available dialogues is shown in Figure \ref{fig:dialogue_list}, in the Appendix. The buttons below this list allow a user to add a blank or formatted dialogue file. Users can also drag and drop files in this screen to upload them. The user is then able to add, delete or edit any particular dialogue. There is also a button to download the whole dataset as a JSON file on this page. Clicking on a dialogue will take users to the individual dialogue annotation screen shown in Figure \ref{fig:turn_list}.

LIDA uses the concept of a ``turn" to organise how a dialogue is displayed and recorded. A turn consists of a query by the user followed by a response from the system, with an unlimited number of labels allowed for each user query. The user query and system response are displayed in the large area on the left of the interface, while the labels for each turn are shown in the scrollable box on the right. There are two forms that these labels can currently take which are particularly relevant for dialogue: multilabel classification and slot-value pair.

An example of multilabel classification is whether the user was informing the system or requesting a piece of information. An example of a slot-value pair is whether the user mentioned the type of restaurant they'd like to eat at (slot: restaurant-type) and if so what it was (value: italian, for example). The front-end code is written in a modular form so that it is easy for researchers using LIDA to add custom types of labels and annotations, such as sequence classification, to LIDA.

\begin{figure*}
  \centering
  \includegraphics[width=0.85\textwidth]{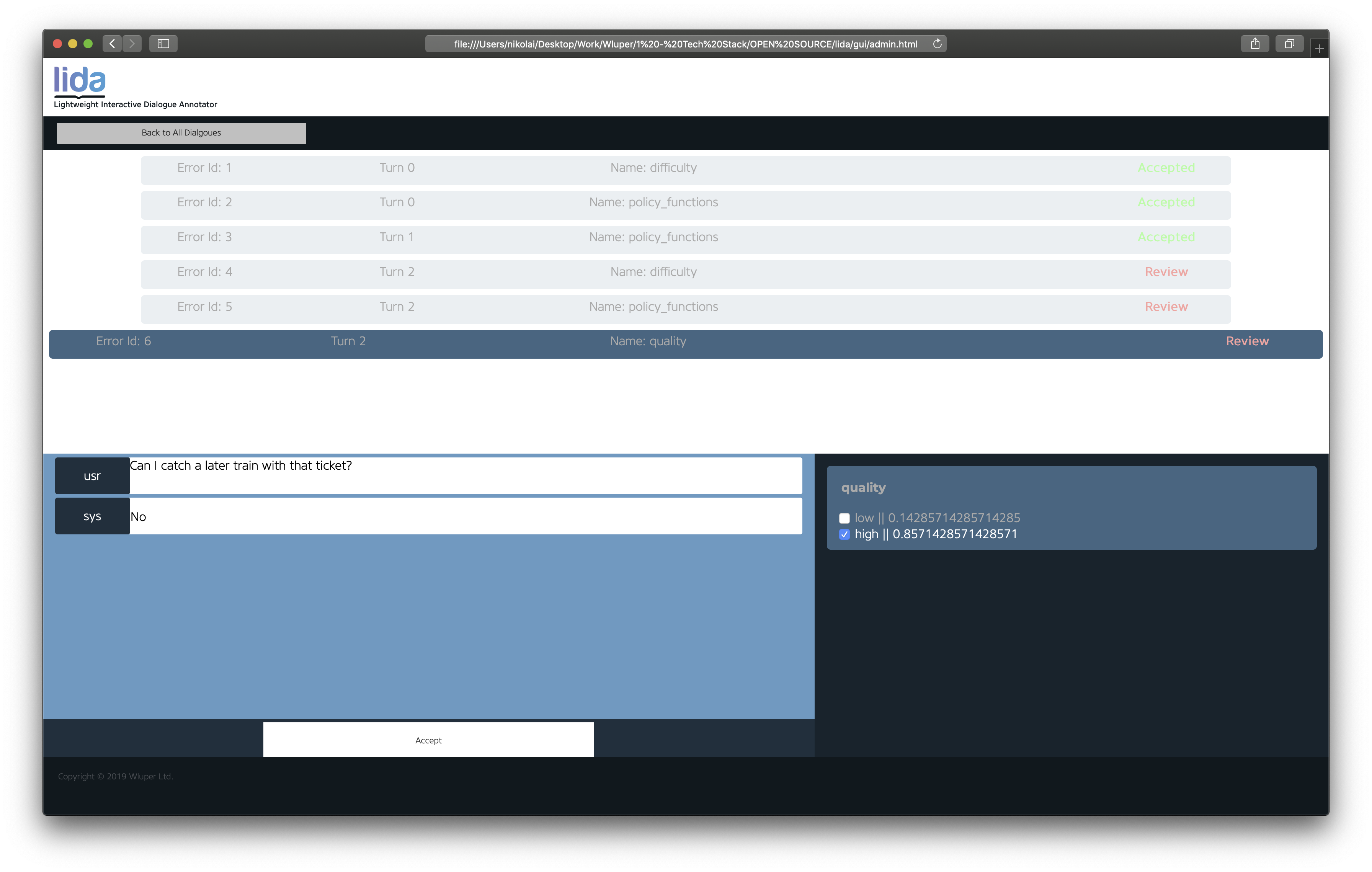}
  \caption{Screenshot of the inter-annotator disagreement resolution screen.}
  \label{fig:interannotator}
\end{figure*}

Once annotation is complete, users can resolve inter-annotator disagreements on the resolution screen. Here, each dialogue is listed along with the number of different people who have annotated it. More annotations can be added by dragging and dropping dialogue files into this screen. When the user clicks on one of these dialogues, they are taken to the resolution screen shown in Figure \ref{fig:interannotator}. Here, all of the disagreements in a dialogue are listed and the label which annotators disagreed on is also shown. The label most frequently selected by annotators is assigned as correct by default, and the arbiter can accept this annotation simply by pressing ``Enter" or else re-label the data item. Once the arbiter has checked an annotation, it is displayed as ``Accepted" in the error list and the dialogue file automatically saved.

\subsection{Use Cases}


\subsubsection{Experimenting with Dialogue Systems}
While generic evaluation metrics are important for understanding the performance of a dialogue system, another important method of evaluation is to talk to the dialogue system and see if it gives subjectively satisfying results. This gives the researcher insight into which part of the system most urgently needs improvement faster than performing more complex error analysis. However, if the user talks to their dialogue system through a terminal interface, they have no way of correcting the system when it answers incorrectly. The researcher should be able to record every interaction they have with their system and correct the predictions of the system easily and quickly. That way, the researcher will be able to use each previous recorded interaction as a test case for future versions of their system.

LIDA is designed with this in mind - a dialogue system can be integrated in the back end so that it will run whenever the user enters a new query in the front end. The user will then be able to evaluate whether the system gave the correct answer and correct the labels it gets wrong using the front end. LIDA will record these corrections and allow the user to download the interaction with their dialogue system with the corrected labels so that it can be used as a test case in future versions of the system. 

\subsubsection{Creating a New Dialogue Dataset} 
Users can create a blank dialogue on LIDA's home screen, then enter queries in the box shown at the bottom of Figure \ref{fig:turn_list}. Along with whole dialogue systems, arbitrary ML models can be added as recommenders in the back end. Once the user hits ``Enter", the query is run through the recommender models in the back end and the suggested annotations displayed for the label. If no recommender is specified in the back end, the label will be left blank. Users can delete turns and navigate between them using ``Enter" or the arrow keys. The name of the dialogue being annotated can be seen next to the ``Back" button at the top left of the screen and can be edited by clicking on it. 

\subsubsection{Annotating An Existing Dataset} 
Datasets can be uploaded via drag-and-drop to the home screen of the system, or paths can be specified in the back end if the system were being used for crowdsourcing. Datasets can be in one of two forms, either a ``.txt" file such as may be produced by a transcription service, or a formatted ``.json" file, a common format for dialogue data \cite{budzianowski2018multiwoz, henderson2014second}. Once the user has uploaded their data, their dialogue(s) will appear on the home screen. The user can click on each dialogue and will be taken to the single dialogue annotation screen shown in Figure \ref{fig:turn_list} to annotate it. If the user uploaded a text file, they will be taken to a dialogue and turn segmentation screen. Following the same constraints imposed in MultiWOZ \cite{budzianowski2018multiwoz} and DSTC \cite{henderson2014second}, this turn segmenter assumes that there are only two participants in the dialogue: the user and the system, and that the user asks the first query. The user separates each utterance in the dialogue by a blank line, and separates dialogues with a triple equals sign (``==="). Once the user clicks ``Done", the text file will automatically be parsed into the correct JSON format and each query run through the recommenders in the back-end to obtain annotation suggestions.

\subsubsection{Resolving Annotator Disagreement} 
Researchers could use LIDA's main interface to crowdsource annotations for a dialogue dataset. Once they have several annotations for each dialogue, they can upload these to the inter-annotator resolution interface of LIDA. The disagreements between annotators will be detected, with a percentage shown beside each label to show how many annotators selected it. The label with the highest percentage of selections is checked by default. The arbiter can accept the majority label simply by pressing ``Enter" and can change errors with the arrow keys to facilitate fast resolution. This interface also displays an averaged (over turns) version of Cohen's Kappa \cite{cohenKappa}, the total number of annotations, total number of errors and averaged (over turns) accuracy.



\subsection{Features\footnote{We refer the reader to visit the public repository for a full documentation \url{https://github.com/Wluper/lida}.} }

\paragraph{Specifying Custom Labels} LIDA's configuration is controlled by a single script in the back end. This script defines which labels will be displayed in the UI and is easy to extend. Users can define their own labels by altering this configuration script. If a user wishes to add a new label, all they need to do is specify the label's name, its type (classification or slot-value pair, currently) and the possible values the classification can take. Alongside the label specification, they can also specify a recommender to use for the label values. The label will then automatically be displayed in the front end. Note that labels in uploaded datasets will only be displayed if the label has an entry in the configuration file.

\paragraph{Custom Recommenders} When creating a dialogue dataset from scratch, LIDA is most powerful when used in conjunction with recommenders which can suggest annotations for user queries to be corrected by the annotator. State-of-the-art tools such as INCEpTION \cite{klie2018inception} emphasise the importance of being able to use recommenders in annotation systems. Users can specify arbitrary ML models to use for each label in LIDA's back end. The back end is written in Python, the de facto language for machine learning, so researchers can directly integrate models written in Python to the back end. This is in contrast to tools such as INCEpTION \cite{klie2018inception} and GATE \cite{cunningham2002gate} which are written in Java and so require extra steps to integrate a Python-based model. To integrate a recommender, the user simply provides an instantiated Python object in the configuration file that has a method called ``transform" that takes a single string and returns a predicted label. 


\paragraph{Dialogue and Turn Segmentation from Raw Data} When uploading a .txt file, users can segment each utterance and each dialogue with a simple interface. This means that raw dialogue data with no labels, such as obtained from a transcription service, can be uploaded and processed into a labelled dialogue. Segmented dialogues and turns are automatically run through every recommender to give suggested labels for each utterance. 



\section{Evaluation}

Table \ref{tab:tool_comparison} shows a comparison of LIDA to other annotation tools. To our knowledge, LIDA is the only annotation tool designed specifically for dialogue systems which supports the full pipeline of dialogue annotation from raw text to labelled dialogue to inter-annotator resolution and can also be used to test the subjective performance of a dialogue system.

To test LIDA's capabilities, we designed a simple experiment: we took a bespoke dataset of 154 dialogues with an average of 3.5 turns per dialogue and a standard deviation of 1.55. The task was to assign three classification labels to each user utterance in each dialogue. Each annotator was given a time limit of 1 hour and told to annotate as many dialogues as they could in that time. We had six annotators perform this task, three of whom were familiar with the system and three of whom had never seen it before.

These annotators annotated an average of 79 dialogues in one hour with a standard deviation of 30, which corresponds to an average of 816.5 individual annotations. The annotators who had never seen the system before annotated an average of 60 dialogues corresponding to an average of 617 individual annotations.

Once we had these six annotations, we performed a second experiment whereby a single arbiter resolved inter-annotator disagreements. In one hour, the arbiter resolved 350 disagreements and noted that resolution was slowest when resolving queries with a high degree of disagreement.

These results show that LIDA provides fast annotation sufficient for collecting large scale data. At the recorded pace of the annotators who had not seen the system before, 100 workers could create a dialogue dataset of 6000 dialogues with approximately $6000 * 3.5 = 21 000$ turns with three annotations per turn in one hour. In large scale collections, such as MultiWOZ \cite{budzianowski2018multiwoz} where 1249 workers were used, much larger datasets with richer annotations could be created. Clearly annotation quantity will depend on the difficulty of the task, length of dialogue and number of labels to be assigned to each utterance but our results suggest that a high speed is achievable. 






\section{Conclusion}
\label{sec:length}

We present LIDA, an open source, web-based annotation system designed specifically for dialogue data. LIDA implements state-of-the-art annotation techniques including recommenders, fully customisable labels and inter-annotator disagreement resolution. LIDA is the only dialogue annotation tool which can handle the full pipeline of dialogue dataset creation from turn and dialogue segmentation to structured conversation data to inter-annotator disagreement resolution. 


Future work will look at adding new label types to LIDA, adding the possibility to have more than two actors in the conversation, a centralised admin page, additional labelling (e.g. co-reference resolution) and in general enhancing usability as users provide feedback.
Our hope is that this work will find applications and usability beyond what we have outlined and developed so far and that with a community effort a modern and highly accessible tool will become widely available.



\bibliography{emnlp-ijcnlp-2019}
\bibliographystyle{acl_natbib}

\newpage

\appendix

\section{Additional System Screenshots \& Scree Cast link:}

Screen Cast Video: \\
\url{https://vimeo.com/329824847}

\clearpage
\begin{figure*}[!htbp]
  \centering
  \includegraphics[width=0.85\textwidth]{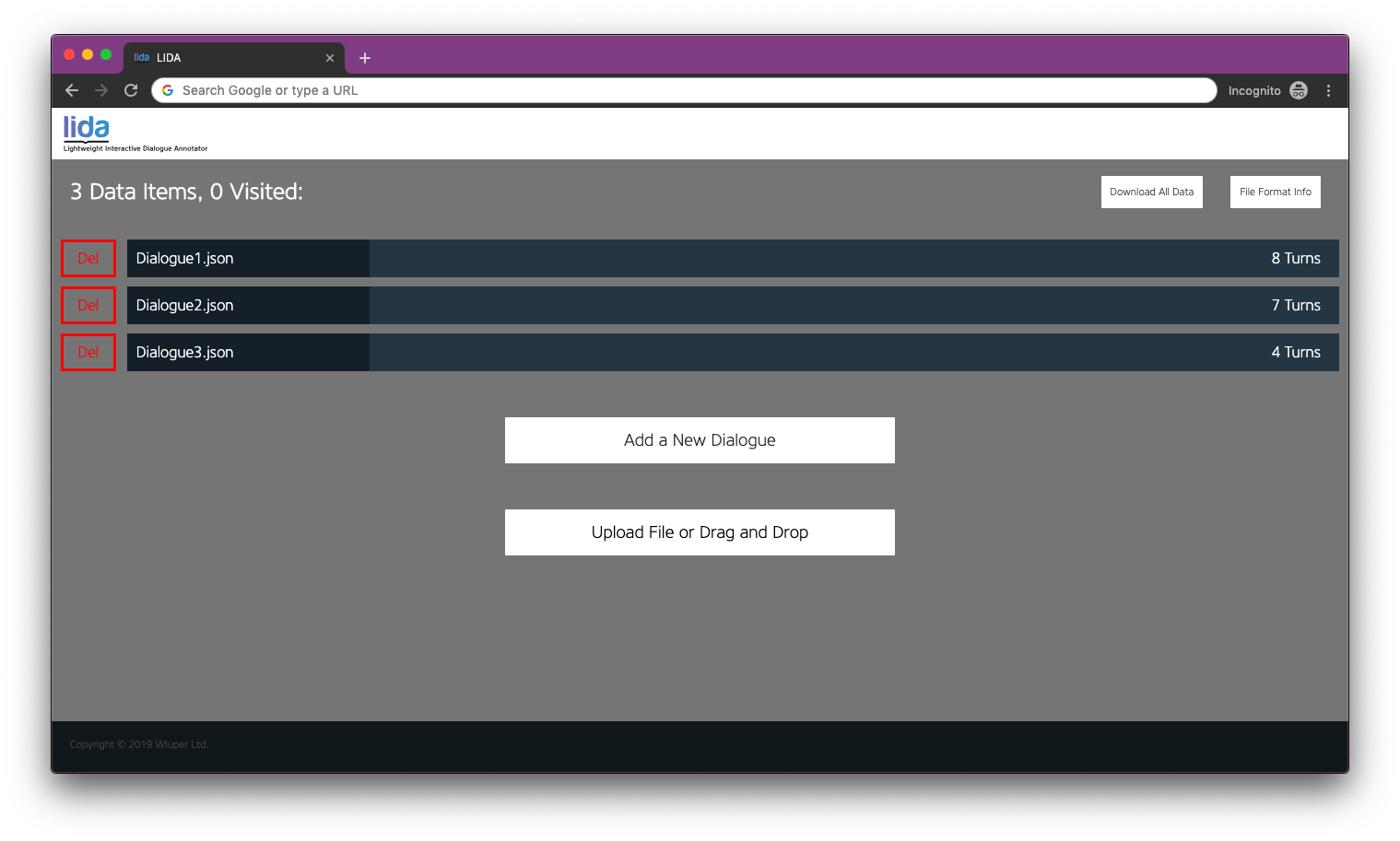}
  \caption{Dialogue List: A list of dialogues are shown on the home page. Users are able to add, delete and edit any dialogue data. The button below allows you to upload a raw text file which leads you to Figure \ref{fig:turn_seg}}
  \label{fig:dialogue_list}
\end{figure*}

\begin{figure*}
  \centering
  \includegraphics[width=0.85\textwidth]{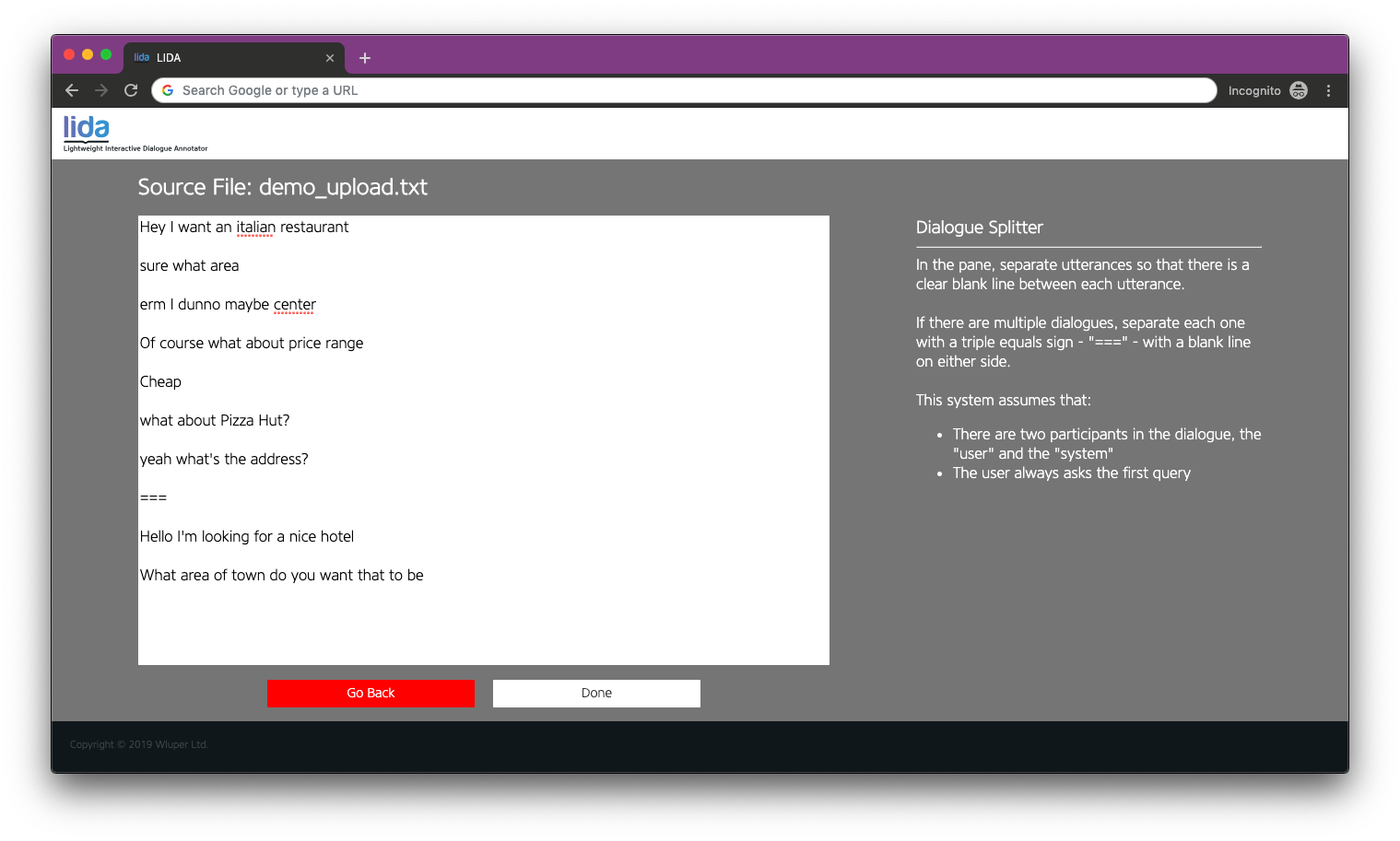}
  \caption{Screenshot of dialogue/turn segmentation: after uploading the raw text, all text is presented in a text area, which allows users to segment them into different dialogues and turns.}
  \label{fig:turn_seg}
\end{figure*}

\begin{figure*}
  \centering
  \includegraphics[width=0.85\textwidth]{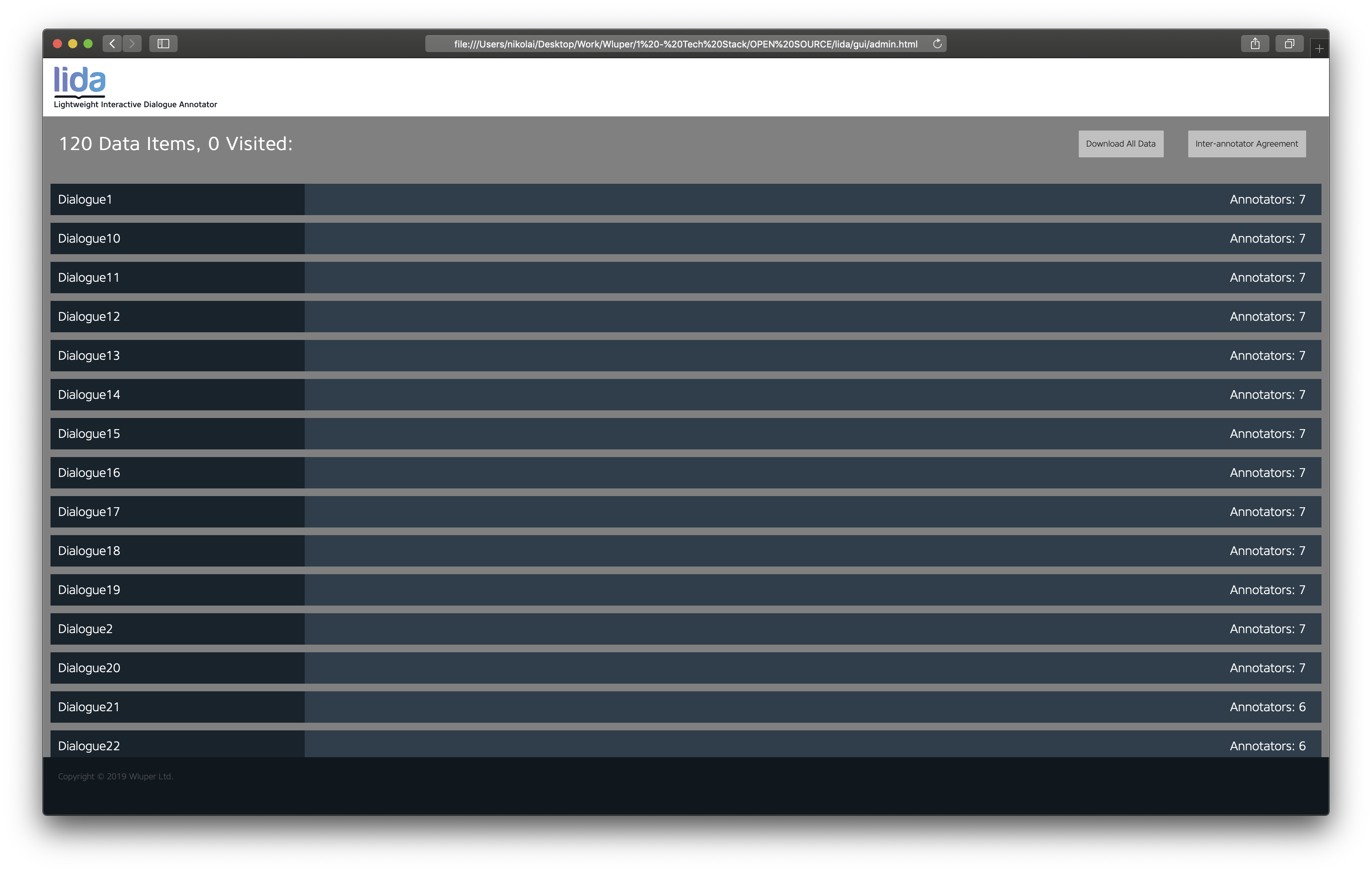}
  \caption{Screenshot of the inter-annotator agreement main dialogue list: clicking on any of the items takes the user to the error resolution screen.}
  \label{fig:interannotator-main-list}
\end{figure*}

\begin{figure*}
  \centering
  \includegraphics[width=0.85\textwidth]{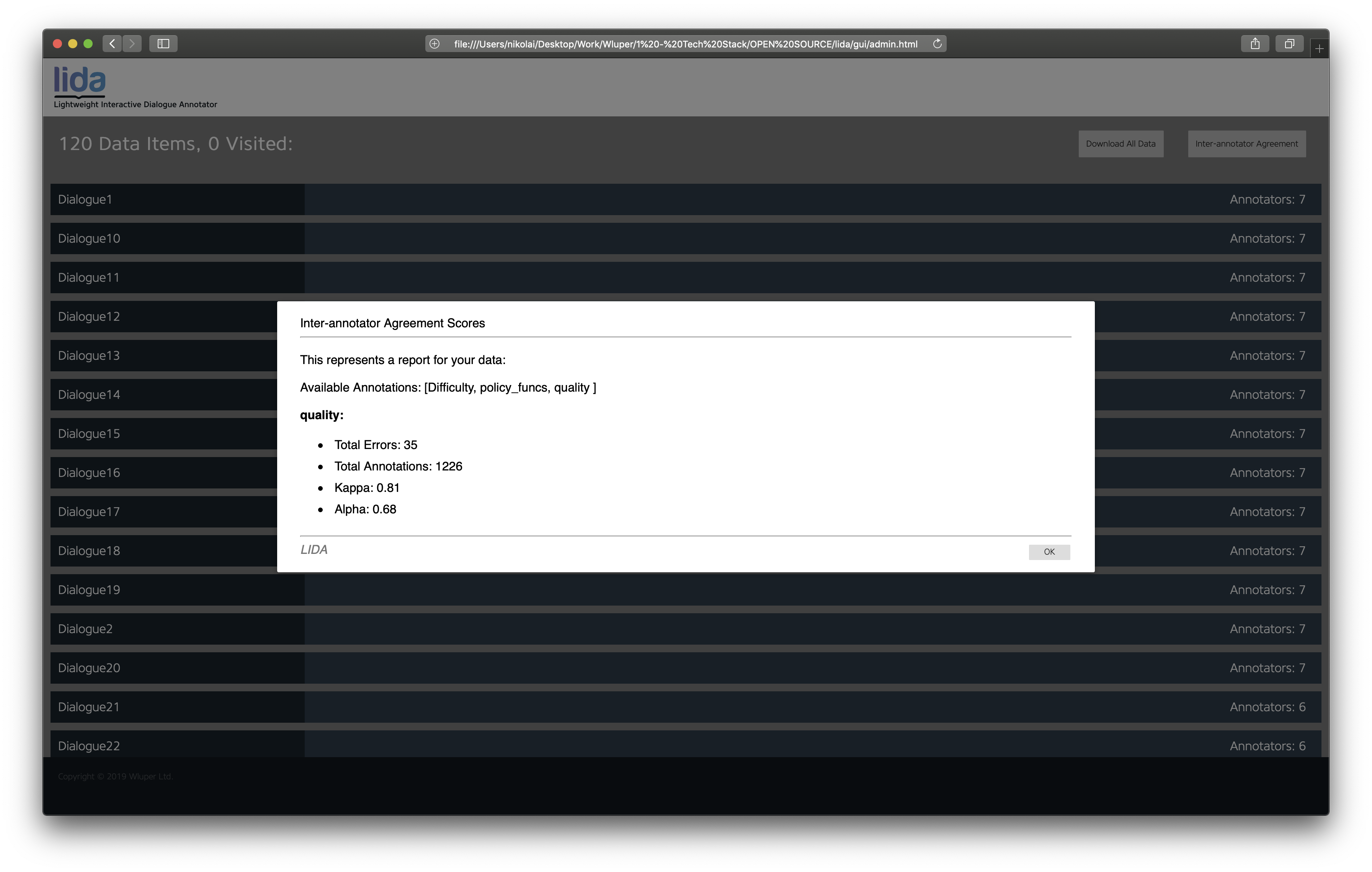}
  \caption{Screenshot of the inter-annotator agreement scores shown by clicking on the ``Inter-annotator Agreement" button in the list screen for inter-annotator disagreement resolution (Figure \ref{fig:interannotator-main-list}).}
  \label{fig:interannotator-scores}
\end{figure*}

\end{document}